\documentclass[11pt]{article}

\PassOptionsToPackage{table}{xcolor}  
\usepackage[final]{acl}
\usepackage[colorinlistoftodos]{todonotes}
\usepackage{times}
\usepackage{latexsym}
\usepackage{url}
\usepackage{multirow} 
\usepackage[T1]{fontenc}

\usepackage[utf8]{inputenc}

\usepackage{microtype}
\usepackage{amsmath}
\usepackage{inconsolata}

\usepackage{graphicx}
\usepackage{booktabs}
\usepackage{array}
\usepackage{amssymb}
\usepackage{color}
\usepackage{fontawesome5}
\usepackage{tcolorbox}    
\usepackage[capitalize,noabbrev,nameinlink]{cleveref}

\title{LongAct: Harnessing Intrinsic Activation Patterns for Long-Context Reinforcement Learning}

\author{
  \textbf{Bowen Ping\textsuperscript{1,*}},
  \textbf{Zijun Chen\textsuperscript{2,*}},
  \textbf{Tingfeng Hui\textsuperscript{3}},
  \textbf{Qize Yu\textsuperscript{1}},
\\
  \textbf{Chenxuan Li\textsuperscript{1}},
  \textbf{Junchi Yan\textsuperscript{2}},
  \textbf{Baobao Chang\textsuperscript{1,\dag}}
\\
\\
  \textsuperscript{1}Peking University,
  \textsuperscript{2}Shanghai Jiao Tong University
\\
  \textsuperscript{3}Beijing University of Posts and Telecommunications
\\
  \textsuperscript{*}Equal contribution \quad \textsuperscript{\dag}Corresponding author
\\
  \small{
    \textbf{Correspondence:} \href{mailto:chbb@pku.edu.cn}{chbb@pku.edu.cn}, \href{mailto:pingbowen23@stu.pku.edu.cn}{pingbowen23@stu.pku.edu.cn}
  }
}
\DeclareUnicodeCharacter{221E}{\ensuremath{\infty}}
\newcolumntype{Z}{>{\centering\arraybackslash}m{1.5cm}}

\begin{document}
\maketitle
\begin{abstract}
Reinforcement Learning (RL) has emerged as a critical driver for enhancing the reasoning capabilities of Large Language Models (LLMs). While recent advancements have focused on reward engineering or data synthesis, few studies exploit the model's intrinsic representation characteristics to guide the training process. In this paper, we first observe the presence of high-magnitude activations within the query and key vectors when processing long contexts.  Drawing inspiration from model quantization—which establishes the criticality of such high-magnitude activations—and the insight that long-context reasoning inherently exhibits a sparse structure, we hypothesize that these weights serve as the pivotal drivers for effective model optimization. Based on this insight, we propose LongAct, a strategy that shifts from uniform to saliency-guided sparse updates. By selectively updating only the weights associated with these significant activations, LongAct achieves an approximate 8\% improvement on LongBench v2 and enhances generalization on the RULER benchmark. Furthermore, our method exhibits remarkable universality, consistently boosting performance across diverse RL algorithms such as GRPO and DAPO. Extensive ablation studies suggest that focusing on these salient features is key to unlocking long-context potential. 
\end{abstract}

\section{Introduction}
Reinforcement Learning (RL) has been proven to be a catalyst for eliciting the reasoning capabilities of Large Language Models (LLMs)~\citep{guo2025deepseek,team2025kimi,hui2025decif}. This capability is particularly pivotal in long-context scenarios. Real-world long-context tasks, such as long-dialogue history understanding and long structured data analysis, are characterized not only by their extensive input lengths but also by the necessity for deep comprehension and complex reasoning over the content. In addition, enhancing long-context understanding is instrumental for LLM agents in managing extended trajectories.


\begin{figure}
    \centering
    \includegraphics[width=1.0\linewidth]{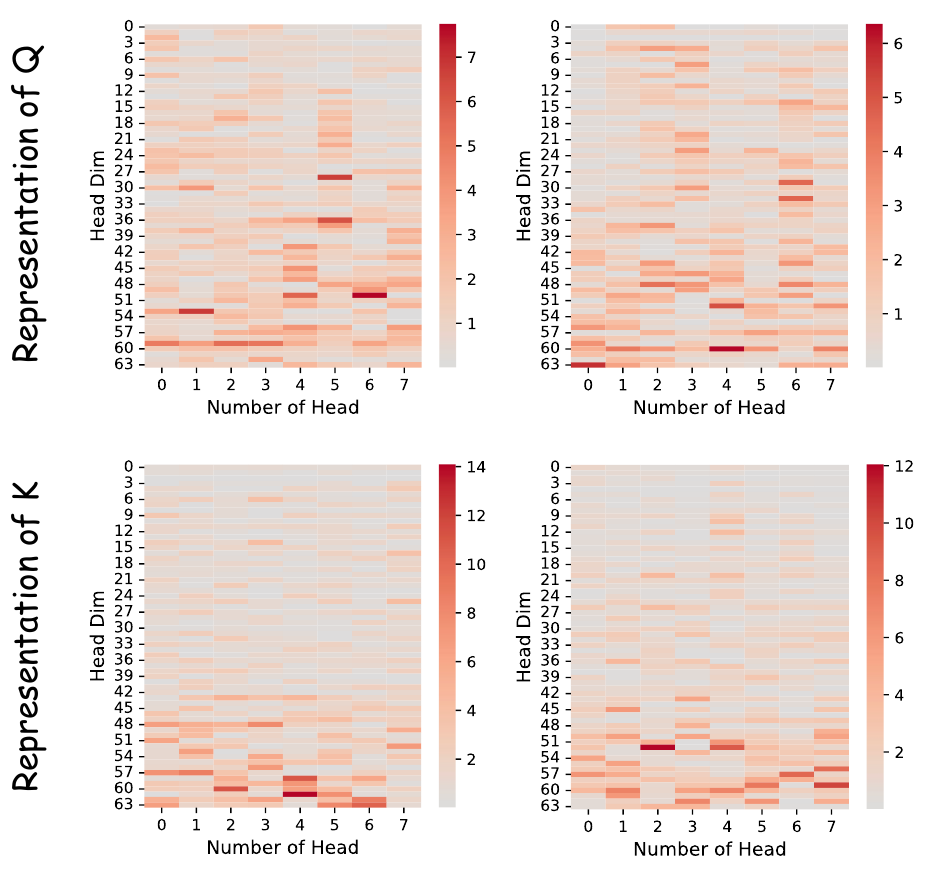}
    \caption{Visualization of the query/key (Q/K) representation magnitudes in Qwen3-8B on the RULER benchmark (Common Words Extraction subset). We show the first 8 attention heads and the first 64 dimensions within each head for clarity. The x-axis denotes the head index and the y-axis denotes the head dimension.}
    \label{fig:1}
\end{figure}

Recently, researchers have begun to explore the application of RL in long-context scenarios~\citep{bai2025longbench,zeng2025glm,ping2025longdpo,ping2026longr}. Existing efforts primarily optimize external supervision signals or training curricula. For instance, some approaches focus on synthesizing high-quality reasoning data~\citep{wang2025loongrl} or employing granular reward functions to mitigate sparse feedback~\citep{longrlvr2025}, while others adopt progressive context scaling strategies~\citep{wan2025qwenlong}. Parallel efforts have even explored modifying the model architecture itself to handle extended sequences~\citep{QwenTeam2025Qwen3Next,team2025kimi_linear}. However, these methods effectively treat the model's internal computation as a black box. Previous work~\citep{hao2024training,deng2025latentreasoningllmsvocabularyspace} suggests that complex deduction relies on continuous "thought trajectories" within the hidden state space rather than merely surface-level token generation. Yet, current long-context RL paradigms largely overlook the features embedded in these latent representations.

To bridge this gap, we propose LongAct, a method that leverages the model's intrinsic activation patterns to guide the training process. Our intuition is grounded in two complementary insights. First, prior studies~\citep{lin2024awq,jin2025massive} demonstrate that hidden dimensions are not equally important—high-magnitude activations often encode disproportionately critical information compared to the rest. Second, long-context inherently exhibits a sparse structure, as evidenced by methods that achieve full-context performance utilizing only a subset of selected tokens~\citep{xiao2024infllm,zhao2025infllm}. We hypothesize that this sparsity extends beyond the token level to the hidden state dimension. As illustrated in \cref{fig:1}, we empirically observe this phenomenon as sparse, high-magnitude activations within the query and key vectors. Identifying these activations as the structural "anchors" for long-context reasoning, LongAct adopts a sparse, saliency-guided strategy, selectively updating only the weights linked to these significant features. This targeted approach yields an approximate 8\% improvement on LongBench v2~\citep{bai2025longbench}, demonstrating that focusing on intrinsic saliency is key to unlocking long-context potential.

LongAct exhibits remarkable universality, enhancing generalization on generic long-context tasks (e.g., RULER and InfiniteBench~\citep{hsieh2024ruler,zhang2024bench})—evidenced by a ~4\% gain on 128K RULER in~\cref{tab:tab2}—while consistently boosting performance across a diverse spectrum of RL algorithms, including GRPO, DAPO, and KL-Cov~\citep{shao2024deepseekmath,yu2025dapo,cui2025entropy} shown in~\cref{tb:ablation_algo}. Furthermore, ablation studies indicate that updating weights associated with high-magnitude activations is critical for these improvements. Specifically, our strategy achieves an overall score of 36.73 on LongBench v2, significantly outperforming methods that update low-magnitude (29.82) shown in~\cref{tb:ablation_massive_values}. Finally, case-level analysis in~\cref{case:massive_values} illustrates that disrupting high-magnitude activations triggers immediate model collapse (e.g., repetitive loops), whereas neutralizing low-magnitude counterparts preserves reasoning coherence.

Our contributions can be summarized as follows:
\begin{itemize}
    \item We propose LongAct that leverages intrinsic high-magnitude activations to guide sparse reinforcement learning.
    
    \item We conduct extensive experiments (using LongBench v2, RULER, etc.) to validate the effectiveness of LongAct. For instance, LongAct achieves an improvement of 8\% on LongBench v2.
    \item We provide in-depth experiments and analysis to elucidate the efficacy of LongAct, identifying high-magnitude activations is critical for model reasoning in long-context scenarios.
\end{itemize}

 
\section{Related work}
\textbf{Reinforcement Learning in Long-context Scenario}
Reinforcement Learning (RL) is widely used in tasks such as mathematics~\citep{yu2025dapo,zheng2025group}. Recently, researchers have begun to explore the application of RL in long-context scenarios. Many researchers modify model architectures, adopting methods such as linear attention and sparse attention~\citep{QwenTeam2025Qwen3Next,team2025kimi_linear,team2025minicpm4,gao2025seerattention} which require pre-training.~\citet{wan2025qwenlong} use progressive context scaling during RL. LongRLVR employs a carefully designed reward function~\citep{longrlvr2025}.~\citet{wang2025loongrl} propose to synthesize better long-context reasoning data. In contrast to prior work, our method leverages the model's internal mechanisms and is complementary to existing approaches.

\textbf{Important Values in Attention Modules} Numerous studies have investigated high-magnitude activation~\citep{dettmers2022gpt3,ahmadian2023intriguing,guo2024active,xu2024slmrec}. Many methods have proven effective; more specifically,~\citet{lin2024awq} preserve weights related to high-magnitude activations with high precision during quantization, while~\citet{liu2024kivi} employ asymmetric quantization guided by the distribution of high-magnitude activations in the KV cache. Several other studies have investigated the impact of RoPE on model activations~\citep{barbero2024round,jin2025massive}. Departing from previous research, we analyze how high-magnitude activations influence performance in long-context reasoning tasks.

\begin{figure*}
    \centering
    \includegraphics[width=0.9\linewidth]{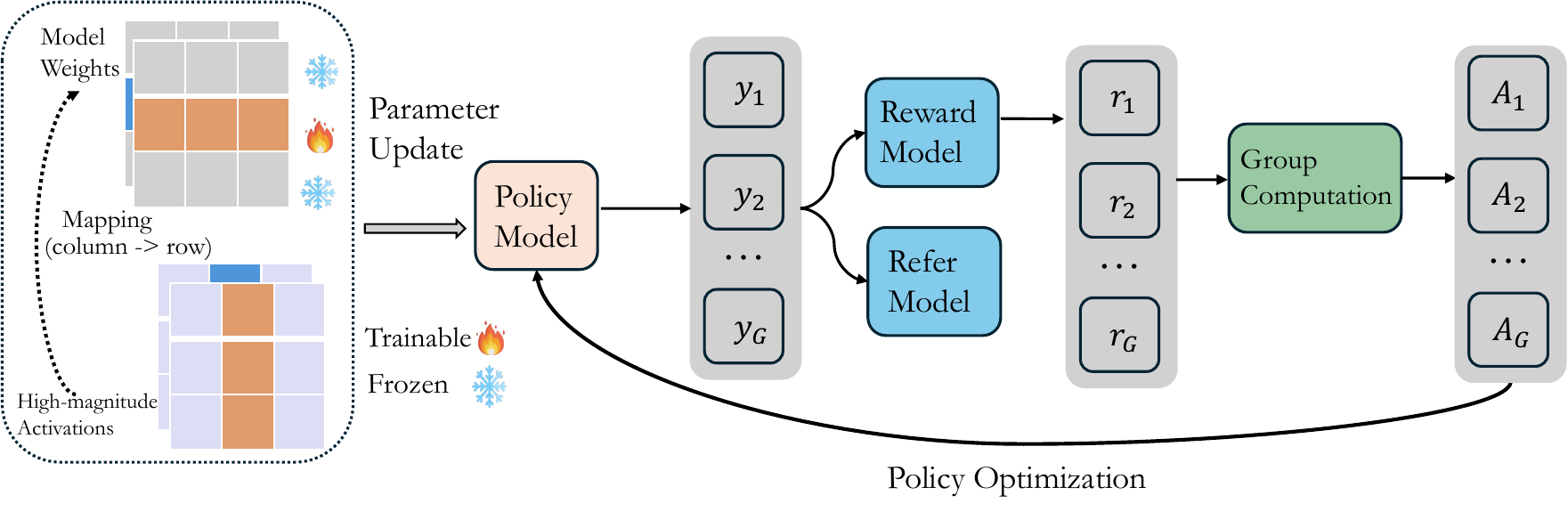}
    \caption{Overview of the LongAct framework. The left panel illustrates the dynamic saliency-guided sparse update mechanism: distinct high-magnitude activations (Orange/Blue columns) in the projections (e.g., Query/Key) dynamically map to their corresponding weight rows for sparse updates, while keeping other parameters frozen. Given the projection weight shape $\mathbf{W} \in \mathbb{R}^{d_{out} \times d_{in}}$, high-magnitude outliers in the activation output channels (columns) correspond directly to specific rows in $\mathbf{W}$. The right panel depicts the standard group-based policy optimization loop.}
    \label{fig:2}
\end{figure*}

\section{Method}
\subsection{Preliminary}
\textbf{High-magnitude Activations}
We focus on the attention layers, which serve as the core components for context modeling. Let $H_Q$ and $H_{KV}$ denote the number of attention heads for the query and key/value branches, respectively.

Let $W_Q \in \mathbb{R}^{H_Q \times D \times d_{\text{model}}}$,  and $W_K \in \mathbb{R}^{H_{KV} \times D \times d_{\text{model}}}$ represent the learnable projection weights, where $D$ is the head dimension. Given an input hidden state $H_{\text{in}} \in \mathbb{R}^{B \times S \times d_{\text{model}}}$, the query and key activations are computed via linear projections:

\begin{equation} \label{eq:1}
    Q = H_{\text{in}} W_Q^\top, \quad K = H_{\text{in}} W_K^\top,
\end{equation}

where $Q \in \mathbb{R}^{B \times S \times H_Q \times D}$ and $K \in \mathbb{R}^{B \times S \times H_{KV} \times D}$, with $B$ denoting the batch size and $S$ the sequence length.

To quantify activation patterns, we compute the $\ell_2$-norm across the sequence dimension. We define the global magnitude matrix $M$ as the expectation over the batch. Since the number of heads may differ for $Q$ and $K$, we denote their magnitude matrices as $M^Q \in \mathbb{R}^{H_Q \times D}$ and $M^K \in \mathbb{R}^{H_{KV} \times D}$, respectively. For the query representation $Q$, the magnitude for head $h$ at feature dimension $d$ is given by:


\begin{equation} \label{eq:2}
    M^Q_{h, d} = \frac{1}{B} \sum_{i=1}^{B} \sqrt{\sum_{s=1}^{S} (Q^{(i)}_{s, h, d})^2}.
\end{equation}

A similar calculation applies to $K$ using $H_{KV}$ heads. As illustrated in~\cref{fig:1}, high-magnitude activations (outliers) consistently appear in specific dimensions. Following insights from quantization~\citep{lin2024awq,jin2025massive}, we identify the specific rows in $W_Q$ and $W_K$ corresponding to these outlier dimensions as the critical parameters for update.

\textbf{Reinforcement Learning in Long-context Scenarios}
The standard reinforcement learning objective in language modeling seeks to optimize an expected reward, regularized by KL divergence.~\citep{schulman2018equivalencepolicygradientssoft}:
\begin{multline}
\max_{\pi_\theta} \mathbb{E}_{x \sim \mathcal{D}, y \sim \pi_{\theta}(\cdot \mid x)} 
\left[ r_{\phi}(x, y) \right] \\
- \beta \mathbb{D}_{\text{KL}} \left[ \pi_{\theta}(y \mid x) \, \middle\| \, \pi_{\text{ref}}(y \mid x) \right]
\end{multline}
where $r_{\phi}(x,y)$ denotes the reward for output $y$ given input $x$ from the policy model $\pi_{\theta}$, and $\pi_{\text{ref}}$ represents the reference model for $\mathbb{D}_{\text{KL}}$ regularization.

Unlike prior works that rely on the parametric knowledge of the policy model $\pi_{\theta}$ to generate an output $y$ from a typically short question $x$, we extend the formulation by incorporating an additional long-context $c$. This requires $\pi_{\theta}$ to first ground relevant information in $c$ before producing reasoning chains to solve $x$:
\begin{multline}
\max_{\pi_\theta} \mathbb{E}_{x, c \sim \mathcal{D}, y \sim \pi_{\theta}(\cdot \mid x, c)} 
\left[ r_{\phi}(x, c, y) \right] \\
- \beta \mathbb{D}_{\text{KL}} \left[ \pi_{\theta}(y \mid x, c) \,\middle\|\, \pi_{\text{ref}}(y \mid x, c) \right]
\end{multline}

Reinforcement Learning with Variable Reward (RLVR) significantly enhances the reasoning capabilities of models. We build our training framework upon Group Relative Policy Optimization (GRPO)~\citep{shao2024deepseekmath}, which eliminates the need for an external critic model by normalizing rewards within a group of outputs. Given a context $c$ and question $x$, the old policy $\pi_{\theta_{\text{old}}}$ generates a group of $G$ outputs $\{y_{i}\}_{i=1}^{G}$, with rewards $\{r_{i}\}_{i=1}^{G}$. The optimization objective is formulated as:
\begin{equation}
\begin{aligned}
\mathcal{J}(\theta) &= \mathbb{E} \Bigg[ \frac{1}{G}\sum_{i=1}^{G} \frac{1}{|y_i|} \sum_{t=1}^{|y_i|} \min \Big( \rho_{i,t} A_{i,t}, \\
& \quad \text{clip}\left(\rho_{i,t}, 1 - \varepsilon, 1 + \varepsilon\right) A_{i,t} \Big) - \beta \mathbb{D}_{\text{KL}} \Bigg]
\end{aligned}
\end{equation}
where the expectation is over $x, c \sim \mathcal{D}$ and $\{y_i\} \sim \pi_{\theta_{\text{old}}}$. Here, $\rho_{i,t} = \frac{\pi_{\theta}(y_{i,t} \mid x, c, y_{i,<t})}{\pi_{\theta_{\text{old}}}(y_{i,t} \mid x, c, y_{i,<t})}$ denotes the probability ratio. Crucially, $A_{i,t}$ is the advantage term derived from group-relative normalization:
\begin{equation}
    A_{i} = \frac{r_i - \text{mean}(\{r_j\}_{j=1}^G)}{\text{std}(\{r_j\}_{j=1}^G)}
\end{equation}
Since recent studies have identified the limitations of naive GRPO~\citep{yu2025dapo,zheng2025group}, we employ DAPO in our implementation for more stable training.

\subsection{Supervised Fine-tuning (Cold Start)}
\label{sec:sft}

The first stage of our pipeline is a Supervised Fine-tuning (SFT) phase, which initializes the base model with a robust policy prior to reinforcement learning. In this stage, we optimize the model parameters $\theta$ by minimizing the standard Cross-Entropy (CE) loss over the gold reasoning trajectories:
\begin{equation}
\mathcal{L}_{\text{SFT}}(\theta) = - \mathbb{E}_{(x, y) \sim \mathcal{D}_{\text{SFT}}} \sum_{t=1}^{|y|} \log \pi_\theta(y_t \mid x, y_{<t}),
\label{eq:sft_loss}
\end{equation}
where $x$ denotes the input context, $y = (y_1, \dots, y_T)$ represents the target chain-of-thought sequence, and $\pi_\theta(y_t \mid x, y_{<t})$ is the probability of the $t$-th token given the context and preceding tokens. This phase ensures the model adapts to the specific output format (e.g., enclosing reasoning processes within \texttt{<think>} tags) required for the subsequent RL stage.

\subsection{LongAct Training Framework}
\label{sec:longrl_framework}

We propose a sparsity-aware training framework that dynamically adapts to the model's intrinsic activation patterns. Our approach consists of a rule-based reward mechanism and a saliency-guided parameter update strategy.

\textbf{Dynamic Saliency-guided Updates.}
The core of LongAct is to restrict gradient updates to the "load-bearing" parameters identified in the Preliminary. We focus on the projection weights $W_Q$ and $W_K$. Taking the Query as an example (the Key follows the same logic):

The query projection weight $W_Q \in \mathbb{R}^{H_Q \times D \times d_{\text{model}}}$ maps the hidden state to the concatenated head outputs. Structurally, $W_Q$ is organized by heads, where the $r$-th \textbf{row} generates the specific feature dimension for a corresponding head. The mapping from a specific head $h \in \{0, \dots, H_Q-1\}$ and its internal dimension $d \in \{0, \dots, D-1\}$ to the global row index $j$ in $W_Q$ is defined as:
\begin{equation}
    j(h, d) = h \cdot D + d.
\end{equation}

At each training step, we utilize the pre-computed global magnitude matrix $M^Q \in \mathbb{R}^{H_Q \times D}$ (Eq.~\ref{eq:2}). Instead of a global top-$k$ selection, we perform intra-head selection to preserve the multi-head structure. For each head $h$, we identify the subset of critical local dimensions $\mathcal{K}_h$:
\begin{equation}
    \mathcal{K}_h = \left\{ d \mid d \in \mathop{\arg\max}_{k} \{ M^Q_{h, d'} \}_{d'=0}^{D-1} \right\},
\end{equation}
where $k = \lfloor \lambda D \rfloor$ is determined by the sparsity ratio $\lambda$ (e.g., 0.3).

We then define a binary gradient mask $\mathbf{G}^Q \in \{0, 1\}^{H_Q \times D \times d_{\text{model}}}$ for the weight matrix $W_Q$. A row $r$ in $W_Q$ is trainable if and only if it corresponds to a selected high-magnitude feature in its respective head:
\begin{equation}
    \mathbf{G}^Q_{r, :} = 
    \begin{cases} 
        \mathbf{1} & \text{if } r \in \{ j(h, d) \mid \forall h, d \in \mathcal{K}_h \} \\
        \mathbf{0} & \text{otherwise}
    \end{cases}
\end{equation}
where $\mathbf{1}$ and $\mathbf{0}$ denote row vectors of size $d_{\text{model}}$. We provide an example in~\cref{app:example}.

During backward propagation, we apply these masks to the gradients:
\begin{equation}
    \nabla W_Q \leftarrow \nabla W_Q \odot \mathbf{G}^Q, \quad \nabla W_K \leftarrow \nabla W_K \odot \mathbf{G}^K.
\end{equation}
The computational overhead of generating the dynamic mask is negligible. The saliency calculation using~\cref{eq:2} and Top-$k$ selection are performed on the collapsed head-dimension tensors ($H \times D$), not the full sequence, rendering the cost minimal compared to the full forward-backward pass. 

It is important to emphasize that the sparse mask is applied \textbf{only to the gradients} of $W_Q$ and $W_K$ during training. All other parameters (i.e., $W_V, W_O$, and MLP layers) receive standard full updates. Detailed visualizations of these activation distributions are provided in \cref{fig_appendix:q,fig_appendix:k,fig_appendix:v}.During inference, the mask is discarded, and the model operates as a standard dense Transformer with zero additional latency or architectural modifications.

\textbf{Reward Formulation.} We employ a rule-based reward function $r(y)$ composed of a format reward and an answer reward:
\begin{equation}
    r(y) = r_{\text{fmt}}(y) + r_{\text{ans}}(y).
\end{equation}
Specifically, $r_{\text{fmt}}(y)$ is set to 1 if the response $y$ contains the tags \texttt{<think>}, \texttt{</think>}, \texttt{<answer>}, and \texttt{</answer>}, and 0 otherwise. $r_{\text{ans}}(y)$ is set to 1 if the answer matches the ground truth, and 0 otherwise.

\begin{table*}[t]
\centering

\resizebox{0.9\linewidth}{!}{
\begin{tabular}{p{5.7cm} | c | c c | c @{\hspace{3pt}} c @{\hspace{3pt}} c}
\toprule

 & & \multicolumn{2}{c|}{\textbf{Difficulty}} & \multicolumn{3}{c}{\textbf{Length}} \\
 
\cmidrule(r){1-2} \cmidrule(lr){3-4} \cmidrule(l){5-7}

\textbf{Model} & \textbf{Overall} & \textbf{Easy} & \textbf{Hard} & \textbf{Short} & \textbf{Medium} & \textbf{Long} \\ 
\midrule

\texttt{Qwen3-8B*}   & 33.60 & 39.58 & 29.90 & 39.44 & 28.37 & 34.26 \\
\texttt{Qwen3-8B-Base}   & 20.68 & 19.27 & 21.54 & 28.33 & 17.67 & 13.89 \\
\texttt{Qwen3-8B-SFT}  & 27.04 & 28.65 & 26.05 & 32.22 & 24.65 & 23.15 \\
\texttt{Qwen3-8B-SFT w/ DAPO} & 32.80 & \textbf{40.10} & 28.30 & 38.33 & 28.37 & 32.41  \\
\texttt{Qwen3-8B-SFT w/ LongAct} & \textbf{36.73} & 38.02 & \textbf{35.93} & \textbf{41.94} & \textbf{33.37} & \textbf{34.72} \\
   \cmidrule(lr){1-1}  \cmidrule(lr){2-2} \cmidrule(lr){3-4} \cmidrule(lr){5-7} 
\texttt{Qwen3-4B*}   & 31.41 & \textbf{34.90} & 29.26 & 35.56 & 27.91 & 31.48 \\
\texttt{Qwen3-4B-Base}   & 17.10 & 19.79 & 15.43 & 21.11 & 13.49 & 17.59 \\
\texttt{Qwen3-4B-SFT}  & 25.65 & 21.35 & 28.30 & 33.33 & 22.33 & 19.44 \\
\texttt{Qwen3-4B-SFT w/ DAPO} & 30.42 & 28.12 & 31.83 & 33.33 & 31.16 & 24.07 \\
\texttt{Qwen3-4B-SFT w/ LongAct} & \textbf{34.24} & 33.07 & \textbf{34.97} & \textbf{37.92} & \textbf{31.63} & \textbf{33.33} \\
\bottomrule
\end{tabular}
}
\caption{Evaluation results (\%) on LongBench v2. Qwen3-8B* and Qwen3-4B* are officially released models available on HuggingFace. Short, Medium, and Long refer to length ranges of <32k, 32-128k, and >128k, respectively. The evaluation is conducted using the official code, and we have set the random seed to 0 to ensure reproducibility.}
\label{tb:exp}
\end{table*}

\begin{table*}[t]
    \centering
    \resizebox{1.0\linewidth}{!}{
        \begin{tabular}{l|ccccc|ccccc}
        \toprule
        \multirow{2}{*}{\textbf{Model}} & \multicolumn{5}{c}{\textbf{Ruler-128k}} & \multicolumn{5}{c}{\textbf{Ruler-64k}} \\
        ~ & \textbf{NIAH-sub}& \textbf{NIAH} & \textbf{VT} & \textbf{QA}& \textbf{Avg} & \textbf{NIAH-sub}& \textbf{NIAH} & \textbf{VT} & \textbf{QA}  & \textbf{Avg}  \\
        \midrule
        \texttt{Qwen3-8B-Base} & 30.63 & 51.88 & 47.68 & 17.60 & 36.95 & 37.93 & 61.76 & 28.08 & 39.70 & 41.87  \\
        \texttt{Qwen3-8B-SFT} & 33.57 & 61.64 & 51.96 & 30.50 & 44.42 & 41.85 & 62.96 & 31.00 & 38.80 & 43.65 \\
        \texttt{Qwen3-8B-SFT w/ DAPO} & 33.92 & 57.64 & \textbf{76.56} & 30.40 & 49.63 & 41.98 & 66.13 & 32.52 & \textbf{40.40} & 45.26  \\
        \texttt{Qwen3-8B-SFT w/ LongAct} &  \textbf{34.55} & \textbf{63.70} & 75.24 & \textbf{31.10} & \textbf{51.15} & \textbf{45.97} & \textbf{69.64} & \textbf{34.96} & 34.90 & \textbf{46.37} \\
        \midrule
        \texttt{Qwen3-4B-Base} & 22.23 & 46.74 & 45.64 & 14.00 & 32.15 & 42.72 & 60.55 & 56.40 & 19.30 & 44.74 \\
        \texttt{Qwen3-4B-SFT} & 26.60 & 53.02 & \textbf{91.44} & 27.20 & 49.57 & 40.92 & 66.20 & 90.92 & 20.20 & 54.56  \\
        \texttt{Qwen3-4B-SFT w/ DAPO} & 31.08 & 54.57 & 90.48 & 28.70 & 51.21 & 42.60 & 67.74 & \textbf{94.96} & 30.90 & 59.05   \\
        \texttt{Qwen3-4B-SFT w/ LongAct} & \textbf{31.50} & \textbf{56.29} & 90.64 & \textbf{32.10} & \textbf{52.63} & \textbf{44.70} & \textbf{69.68} & 94.52 & \textbf{36.30} & \textbf{61.30}  \\
        \bottomrule
        \end{tabular}
    }
    \caption{Evaluation results (\%) on Ruler-128K and Ruler-64K. NIAH-sub is derived from the multi-key level 2, multi-key level 3 and multi-value tasks from NIAH.}
    \label{tab:tab2}
\end{table*}

\section{Experimental Results}
\subsection{Set Up}
\textbf{Training Setting}
We use Qwen3-8B-Base~\footnote{\url{https://huggingface.co/Qwen/Qwen3-8B-Base}} as backbone. All experiments are conducted on 8 NVIDIA H800 80GB GPUs. 
\begin{itemize}
    \item \textbf{Supervised Fine-tuning} We conduct our training using LLaMA-Factory~\footnote{\url{https://github.com/hiyouga/LLaMA-Factory}}. For the cold-start phase, we utilize 20k open-source instruction-following samples from AM-DeepSeek-R1-0528-Distilled~\footnote{\url{https://huggingface.co/datasets/a-m-team/AM-DeepSeek-R1-0528-Distilled}} and we employ a cosine learning rate scheduler with a peak rate of 2e-5 and a warmup ratio of 0.03. We set the maximum sequence length to 16,384, with sequence packing and training for 900 steps.
    \item \textbf{Reinforcement Learning} We use verl~\footnote{\url{https://github.com/volcengine/verl}} for RL and use DAPO~\citep{yu2025dapo}. We curate a training dataset by mixing DocQA-RL-1.6K~\footnote{\url{https://huggingface.co/datasets/Tongyi-Zhiwen/DocQA-RL-1.6K}} and randomly 1K samples from MemAgent~\citep{yu2025memagent}. We set the learning rate to 1e-6, with a maximum sequence length of 32,768, a batch size of 8, a rollout number of 16, a temperature of 1.0, and a maximum output length of 4096 for sampling. Unless otherwise specified, we set the default sparsity ratio $\lambda$ for LongAct to \textbf{0.3} (i.e., updating only the top 30\% of high-magnitude weights) based on our ablation studies.
    
\end{itemize}

\textbf{Evaluation Setting} We employ a comprehensive suite of long-context benchmarks to rigorously evaluate model performance:
\begin{itemize}
    \item \textbf{LongBench v2}~\citep{bai2025longbench}: A challenging benchmark focused on realistic long-context deeper understanding. It employs a multiple-choice question format to facilitate rigorous and objective evaluation.
    
    \item \textbf{RULER}~\citep{hsieh2024ruler}: A benchmark designed to evaluate the effective context window size through diverse synthetic tasks.
    
    \item \textbf{InfiniteBench}~\citep{zhang2024bench}: A dataset targeting diverse long-context capabilities, covering heterogeneous tasks such as retrieval (Re.Pa, Re.Nu), summarization (En.Sum), QA (En.QA, Zh.QA), and multi-choice reasoning (En.MC).
\end{itemize}
For inference, we utilize vLLM~\footnote{\url{https://github.com/vllm-project/vllm}} with a decoding temperature of 1.0. All evaluations are conducted using official scripts with a fixed random seed of 0 to ensure reproducibility.

\textbf{Baselines} We conduct comparisons using the \texttt{Qwen3} family at both 4B and 8B scales. The baselines include: (1) the officially released \texttt{Qwen3-Base} models; (2) our reproduced \texttt{Qwen3-SFT} models obtained from the Cold Start phase; and (3) the \texttt{Qwen3-SFT w/ DAPO} variants, which undergo standard full-parameter RL training.

\begin{table*}[t]
\centering
\begin{tabular}{l|cccccc|c}
    \toprule
    \bf Methods & \bf Re.Pa & \bf Re.Nu & \bf En.Sum & \bf En.QA & \bf Zh.QA & \bf En.MC  & \bf Avg. \\
    \midrule
    \texttt{Qwen3-8B-SFT}  & 85.76 & \textbf{86.44} & 14.18 & 25.09 & 14.47 & 54.15 & 46.68 \\
    \texttt{Qwen3-8B-SFT w/ DAPO}  & 85.93 & 83.56 & 11.86 & \textbf{27.45} & \textbf{23.54} & 56.33 & 48.11 \\
    \texttt{Qwen3-8B-SFT w/ LongAct}  & \textbf{86.44} & 85.76 & \textbf{15.83} & 27.33 & 21.61 & \textbf{59.39} & \textbf{49.39} \\

    \midrule
    \texttt{Qwen3-4B-SFT}  & 86.78  & 83.44 & 12.76 & 12.68 & 9.65 & 53.28 & 43.10 \\
    \texttt{Qwen3-4B-SFT w/ DAPO}  & \textbf{87.63} & 84.75 & \textbf{17.78} & 14.83  & 10.39 & 54.59 & 44.99  \\
    \texttt{Qwen3-4B-SFT w/ LongAct}  & 86.61 & \textbf{86.95} & 15.57 & \textbf{21.06} & \textbf{11.61}  & \textbf{56.33}  & \textbf{46.36}  \\
    \bottomrule
\end{tabular}
\caption{Evaluation results (\%) on InfiniteBench. }
\label{tab:tab3}
\end{table*}

\begin{table*}[t]
\centering
\resizebox{0.9\linewidth}{!}{
\begin{tabular}{p{5.7cm} | c | c c | c @{\hspace{3pt}} c @{\hspace{3pt}} c}
\toprule
 & & \multicolumn{2}{c|}{\textbf{Difficulty}} & \multicolumn{3}{c}{\textbf{Length}} \\
\cmidrule(r){1-2} \cmidrule(lr){3-4} \cmidrule(l){5-7}
\textbf{Model} & \textbf{Overall} & \textbf{Easy} & \textbf{Hard} & \textbf{Short} & \textbf{Medium} & \textbf{Long} \\
\midrule

\texttt{Qwen3-8B-SFT} & 27.04 & 28.65 & 26.05 & 32.22 & 24.65 & 23.15 \\
\hspace{1.0em}\texttt{+ DAPO}      & \textbf{36.73} & 38.02 & \textbf{35.93} & \textbf{41.94} & 33.37 & \textbf{34.72} \\
\hspace{1.0em}\texttt{+ GRPO}      & 35.04 & \textbf{41.15} & 31.27 & 37.92 & \textbf{36.05} & 28.24 \\
\hspace{1.0em}\texttt{+ CLIP-conv} & 35.04 & 40.76 & 31.51 & 37.08 & 34.42 & 32.87 \\
\hspace{1.0em}\texttt{+ KL-conv}   & 34.24 & 37.37 & 32.32 & 40.00 & 32.44 & 28.24 \\

\cmidrule(lr){1-1} \cmidrule(lr){2-2} \cmidrule(lr){3-4} \cmidrule(lr){5-7}

\texttt{Qwen3-4B-SFT} & 25.65 & 21.35 & 28.30 & 33.33 & 22.33 & 19.44 \\
\hspace{1.0em}\texttt{+ DAPO}      & \textbf{34.24} & 33.07 & 34.97 & \textbf{37.92} & 31.63 & 33.33 \\
\hspace{1.0em}\texttt{+ GRPO}      & 33.45 & \textbf{34.64} & 32.72 & 37.64 & 28.49 & 36.34 \\
\hspace{1.0em}\texttt{+ CLIP-conv} & 32.11 & 32.55 & 31.83 & 33.06 & 30.00 & 34.72 \\
\hspace{1.0em}\texttt{+ KL-conv}   & 33.90 & 32.03 & \textbf{35.05} & 32.92 & \textbf{33.02} & \textbf{37.27} \\

\bottomrule
\end{tabular}
}
\caption{Evaluation results (\%) on LongBench v2 under different RL algorithms.}
\label{tb:ablation_algo}
\end{table*}

\begin{table*}[t]
\centering
\resizebox{0.9\linewidth}{!}{
\begin{tabular}{p{5.7cm} | c | c c | c @{\hspace{3pt}} c @{\hspace{3pt}} c}
\toprule
 & & \multicolumn{2}{c|}{\textbf{Difficulty}} & \multicolumn{3}{c}{\textbf{Length}} \\
\cmidrule(r){1-2} \cmidrule(lr){3-4} \cmidrule(l){5-7}
\textbf{Model} & \textbf{Overall} & \textbf{Easy} & \textbf{Hard} & \textbf{Short} & \textbf{Medium} & \textbf{Long} \\
\midrule

\texttt{Qwen3-8B-SFT} & 27.04 & 28.65 & 26.05 & 32.22 & 24.65 & 23.15 \\
\hspace{1.0em}\texttt{+ random}         & 28.63 & 30.21 & 27.65 & 28.33 & 30.70 & 25.00 \\
\hspace{1.0em}\texttt{+ min values}     & 29.82 & 32.81 & 27.97 & 35.00 & 26.98 & 26.85 \\
\hspace{1.0em}\texttt{+ massive values} & \textbf{36.73} & \textbf{38.02} & \textbf{35.93} & \textbf{41.94} & \textbf{33.37} & \textbf{34.72} \\

\cmidrule(lr){1-1} \cmidrule(lr){2-2} \cmidrule(lr){3-4} \cmidrule(lr){5-7}

\texttt{Qwen3-4B-SFT} & 27.04 & 28.65 & 26.05 & 32.22 & 24.65 & 23.15 \\
\hspace{1.0em}\texttt{+ random}         & 29.03 & 30.21 & 28.30 & 30.00 & 28.84 & 27.78 \\
\hspace{1.0em}\texttt{+ min values}     & 30.22 & 23.96 & 34.08 & 27.78 & 31.16 & 32.41 \\
\hspace{1.0em}\texttt{+ massive values} & \textbf{34.24} & \textbf{33.07} & \textbf{34.97} & \textbf{37.92} & \textbf{31.63} & \textbf{33.33} \\

\bottomrule
\end{tabular}
}
\caption{Evaluation results (\%) on LongBench v2. Ablation on activation selection strategies.}
\label{tb:ablation_massive_values}
\end{table*}

\subsection{Main Results}
\paragraph{Results on LongBench v2.}
~\cref{tb:exp} summarizes our LongBench v2 results. Overall, \texttt{Qwen3-8B-SFT w/ LongAct} achieves the best performance at \textbf{36.73}, improving by \textbf{+3.93} over \texttt{Qwen3-8B-SFT w/ DAPO} (32.80) and \textbf{+9.69} over the cold-start \texttt{Qwen3-8B-SFT} model (27.04). Notably, our method also outperforms the officially released \texttt{Qwen3-8B*} by \textbf{+3.13} (33.60 $\rightarrow$ 36.73), indicating that \texttt{LongAct} provides gains beyond standard post-training recipes.
We observe a consistent trend on the smaller model: \texttt{Qwen3-4B-SFT w/ LongAct} reaches \textbf{34.24}, yielding \textbf{+3.82} over \texttt{Qwen3-4B-SFT w/ DAPO} (30.42) and surpassing \texttt{Qwen3-4B*} by \textbf{+2.83} (31.41 $\rightarrow$ 34.24). These results suggest that our approach scales robustly across model sizes.

\paragraph{Difficulty Breakdown.}
\texttt{LongAct} excels on \emph{hard} instances, which are highly sensitive to error accumulation. On the 8B model, it achieves \textbf{35.93}, substantially surpassing \texttt{SFT w/ DAPO} (\textbf{+7.63}) and \texttt{SFT} (\textbf{+9.88}). Similarly, the 4B model reaches \textbf{34.97}, outperforming all baselines. While \texttt{DAPO} performs well on easy splits, \texttt{LongAct} offers a more balanced profile, significantly lifting performance on hard tasks without compromising overall capability.



\paragraph{Length Breakdown.}
We further analyze robustness across input lengths. \texttt{LongAct} achieves the best results across \emph{short/medium/long} categories. For 8B model, \texttt{LongAct} attains \textbf{41.94} (Short), \textbf{33.37} (Medium), and \textbf{34.72} (Long), with particularly strong gains compared to \texttt{SFT w/ DAPO} across input lengths (38.33 $\rightarrow$ 41.94 (\textbf{+3.61}), 28.37 $\rightarrow$ 33.37 (\textbf{+5.00}), 32.41 $\rightarrow$ 34.72 (\textbf{+2.31}) for Short, Medium and Long respectively). 
On 4B model, \texttt{LongAct} significantly improves the long split to \textbf{33.33}, which is \textbf{+9.26} over \texttt{SFT w/ DAPO} (24.07) and \textbf{+1.85} over \texttt{Qwen3-4B*} (31.48). These improvements indicate that \texttt{LongAct} enhances stability as context length increases, rather than overfitting to shorter-context behaviors.

\subsection{Ablation on More Long-context Benchmarks}
\paragraph{Results on RULER-128K and RULER-64K.}
\cref{tab:tab2} reports ablations on RULER under two context lengths (128K and 64K).
Overall, \texttt{LongAct} consistently improves long-context performance across both model sizes. For 8B model, \texttt{Qwen3-8B-SFT w/ LongAct} achieves the best average score on RULER-128K (\textbf{51.15}), improving over \texttt{Qwen3-8B-SFT} (44.42) and \texttt{Qwen3-8B-SFT w/ DAPO} (49.63) by \textbf{+6.73} and \textbf{+1.52}, respectively. 
On RULER-64K, \texttt{LongAct} also yields the highest performance with an average of \textbf{46.37}, outperforming \texttt{Qwen3-8B-SFT} (43.65) and \texttt{Qwen3-8B-SFT w/ DAPO} (45.26). 
We observe the same trend for the 4B model, where \texttt{LongAct} delivers the best overall averages at both 128K and 64K, suggesting that the gains are robust and scale across model sizes.

\paragraph{Results on InfiniteBench.}
\cref{tab:tab3} summarizes results on InfiniteBench.
\texttt{LongAct} achieves the best average performance for both model sizes.
For 8B, \texttt{Qwen3-8B-SFT w/ LongAct} attains the highest average score (\textbf{49.39}), improving over \texttt{Qwen3-8B-SFT} (46.68) by \textbf{+2.71} and over \texttt{Qwen3-8B-SFT w/ DAPO} (48.11) by \textbf{+1.28}. 
The gains are driven by consistent improvements on En.Sum (14.18 $\rightarrow$ 15.83) and En.MC (54.15 $\rightarrow$ {59.39}), while remaining competitive on retrieval and QA tasks (e.g., Re.Pa: 85.76 $\rightarrow$ {86.44}).
For 4B, \texttt{Qwen3-4B-SFT w/ LongAct} also yields the best average (\textbf{46.36}), outperforming \texttt{Qwen3-4B-SFT} (43.10) and \texttt{Qwen3-4B-SFT w/ DAPO} (44.99) by \textbf{+3.26} and \textbf{+1.37}, respectively.
Notably, \texttt{LongAct} substantially boosts long-context QA for smaller models (En.QA: 12.68 $\rightarrow$ {21.06}; Zh.QA: 9.65 $\rightarrow$ {11.61}) and improves En.MC (53.28 $\rightarrow$ {56.33}).
Overall, these results indicate that \texttt{LongAct} provides consistent and general improvements across heterogeneous long-context tasks.


\subsection{Ablation on More Reinforcement Learning Methods}
\cref{tb:ablation_algo} evaluates our training recipe under different RL algorithms on top of the cold-start \textit{SFT} models.
Across both 4B and 8B models, applying RL consistently improves long-context performance, and our method maintains stable gains under all RL algorithms, including \texttt{DAPO}, \texttt{GRPO}, \texttt{CLIP-conv}, and \texttt{KL-conv} on LongBench v2.
Among them, \texttt{DAPO} achieves the best overall results on both 8B and 4B and yields the most balanced improvements across difficulty and length splits, while other algorithms remain competitive but exhibit stronger trade-offs (e.g., favoring easier or medium-length subsets over the longest-context subset).
These results suggest that the effectiveness of our method does not rely on a particular RL algorithm and is generalizable.
Unless otherwise specified, we adopt \texttt{DAPO} as the default in subsequent experiments.

\subsection{Ablation on Selecting Activations}
\label{sec:ablation_activation}
~\cref{tb:ablation_massive_values} evaluates different activation selection strategies during training.
Across both backbones, selecting \textit{massive values} consistently yields the best performance, delivering large gains over the SFT baseline and clearly outperforming \textit{random} selection and \textit{min values}.
In contrast, \textit{random} and \textit{min values} provide only modest improvements and are less consistent across difficulty and length splits.
These results suggest that focusing updates on salient (high-magnitude) activations is crucial for effectively improving long-context capability.

\subsection{Ablation on the Sparsity in RL Parameter Updates}
~\cref{tb:ablation_sparsity} studies how the update sparsity (i.e., the percentage of selected massive values) affects performance.
We observe that using a moderate sparsity consistently yields the best overall results.
For both 8B and 4B backbones, selecting 30\% massive values achieves the highest overall score and the most balanced improvements across difficulty and length splits, especially on the \textit{Hard} and \textit{Long} subsets.
Increasing the ratio to 40\% does not further improve overall performance and can introduce regressions on some partitions, suggesting that overly dense updates may weaken the benefit of targeting the most salient parameters.
Unless otherwise specified, we set the default sparsity ratio to 30\% in subsequent experiments.

\subsection{Short-Context Generalization}
To verify that LongAct is not specialized only for long-context tasks, we evaluate on several standard short-context benchmarks. As shown in \cref{tab:short_context}, LongAct consistently outperforms both the SFT baseline and full-parameter DAPO across GSM8K, HumanEval, and TruthfulQA, suggesting that the saliency-guided sparse update strategy serves as a general robust RL optimizer rather than a long-context-specific technique.

\begin{table}[t]
\centering
\resizebox{0.95\linewidth}{!}{
\begin{tabular}{lccc}
\toprule
\textbf{Method} & \textbf{GSM8K} & \textbf{HumanEval} & \textbf{TruthfulQA} \\
\midrule
Qwen3-8B-SFT       & 71.94 & 68.90 & 67.76 \\
\quad w/ DAPO       & 78.08 & 69.51 & 68.95 \\
\quad w/ LongAct    & \textbf{80.13} & \textbf{73.17} & \textbf{69.52} \\
\bottomrule
\end{tabular}
}
\caption{Evaluation results (\%) on short-context benchmarks. LongAct consistently outperforms both SFT and full-parameter DAPO.}
\label{tab:short_context}
\end{table}

\section{Analysis}

To investigate the underlying mechanism of long-context capabilities, we conduct a perturbation analysis on Qwen3-8B using real-world cases from LongBench v2. As illustrated in~\cref{case:massive_values}, to be specific, we isolate the impact of activation magnitude by selectively disrupting the top 30\% ("Qwen3-8B w/o High-magnitude Activations") versus the bottom 30\% ("Qwen3-8B w/o Non High-magnitude Activations") of activations. Specifically, the selected activations are clamped to the global mean value, calculated by averaging the entire query (or key) tensor across all heads, sequence lengths, and feature dimensions.

Comparing these responses, we observe that "Qwen3-8B w/o Non High-magnitude Activations" retains its logical coherence. As shown in~\cref{case:massive_values}, the generated CoT maintains a structured flow—correctly utilizing logical connectors like "Alternatively" and "However"—and successfully derives the correct answer. This suggests that the core reasoning process is less dependent on these ``quiet'' activations. Conversely, disrupting the top 30\% of high-magnitude activations leads to immediate model collapse. The output degrades into repetitive loops (e.g., the "3333..." pattern visible in~\cref{case:massive_values}).

The stark contrast between these outcomes underscores that high-magnitude activations are pivotal for maintaining reasoning in long-context scenarios. This observation corroborates prior findings~\citep{lin2024awq,liu2024kivi,jin2025massive}. Crucially, our empirical results extend this insight, demonstrating that leveraging these critical activations to guide the training process is highly effective for enhancing long-context capabilities.

\section{Conclusion}
We propose LongAct, a robust method that leverages the model's internal representations to enhance long-context reasoning.

\section*{Limitations}
Limited by computing resources, we could not use larger models for reinforcement learning. We will explore the scaling effects of our approach in future work.

\section*{Acknowledgments}
This work is supported by the National Science Foundation of China under Grant No.61876004.


\appendix

\section{Appendix}

\paragraph{Illustrative Example.}
\label{app:example}
To clarify the indexing mechanism, consider a simplified configuration with $H=2$ heads, head dimension $D=4$, and a selection ratio $\lambda=0.3$, resulting in $k = \lfloor 0.3 \times 4 \rfloor = 1$ active dimension per head.
Suppose the computed importance scores for the two heads are:
\begin{align*}
    \mathbf{M}_{0} &= [0.8, 0.2, \mathbf{0.9}, 0.5] \quad \rightarrow \quad \text{idx}_0 = 2 \\
    \mathbf{M}_{1} &= [0.3, \mathbf{0.7}, 0.6, 0.4] \quad \rightarrow \quad \text{idx}_1 = 1
\end{align*}
The system then maps these local head indices to the global row indices of the projection weight $\mathbf{W} \in \mathbb{R}^{(H \cdot D) \times d_{in}}$:
\begin{itemize}
    \item \textbf{Head 0:} Global Row $g_0 = 0 \times 4 + 2 = \mathbf{2}$.
    \item \textbf{Head 1:} Global Row $g_1 = 1 \times 4 + 1 = \mathbf{5}$.
\end{itemize}
Consequently, only rows $\{2, 5\}$ of $\mathbf{W}$ (2 out of 8 total rows) are updated, while the rest remain frozen.

\begin{table*}[t]
\centering
\resizebox{0.9\linewidth}{!}{
\begin{tabular}{p{5.7cm} | c | c c | c @{\hspace{3pt}} c @{\hspace{3pt}} c}
\toprule
 & & \multicolumn{2}{c|}{\textbf{Difficulty}} & \multicolumn{3}{c}{\textbf{Length}} \\
\cmidrule(r){1-2} \cmidrule(lr){3-4} \cmidrule(l){5-7}
\textbf{Model} & \textbf{Overall} & \textbf{Easy} & \textbf{Hard} & \textbf{Short} & \textbf{Medium} & \textbf{Long} \\
\midrule

\texttt{Qwen3-8B-SFT} & 27.04 & 28.65 & 26.05 & 32.22 & 24.65 & 23.15 \\
\hspace{1.0em}\texttt{+ 20\% massive values} & 32.41 & 39.06 & 28.30 & 39.44 & 30.23 & 25.00 \\
\hspace{1.0em}\texttt{+ 30\% massive values} & \textbf{36.73} & 38.02 & \textbf{35.93} & \textbf{41.94} & 33.37 & \textbf{34.72} \\
\hspace{1.0em}\texttt{+ 40\% massive values} & 35.98 & \textbf{40.62} & 33.12 & 38.33 & \textbf{34.88} & 34.26 \\

\cmidrule(lr){1-1} \cmidrule(lr){2-2} \cmidrule(lr){3-4} \cmidrule(lr){5-7}

\texttt{Qwen3-4B-SFT} & 27.04 & 28.65 & 26.05 & 32.22 & 24.65 & 23.15 \\
\hspace{1.0em}\texttt{+ 20\% massive values} & 30.42 & 30.21 & 30.55 & 29.44 & 30.23 & 32.41 \\
\hspace{1.0em}\texttt{+ 30\% massive values} & \textbf{34.24} & \textbf{33.07} & \textbf{34.97} & 37.92 & \textbf{31.63} & \textbf{33.33} \\
\hspace{1.0em}\texttt{+ 40\% massive values} & 32.80 & 31.77 & 33.44 & \textbf{41.11} & 28.37 & 27.78 \\

\bottomrule
\end{tabular}
}
\caption{Ablation on sparsity (percentage of selected massive values). Results are reported in \%. Bold indicates the best value within each model (8B / 4B) for each column.}
\label{tb:ablation_sparsity}
\end{table*}

\begin{figure*}
    \centering
    \includegraphics[width=1.0\linewidth]{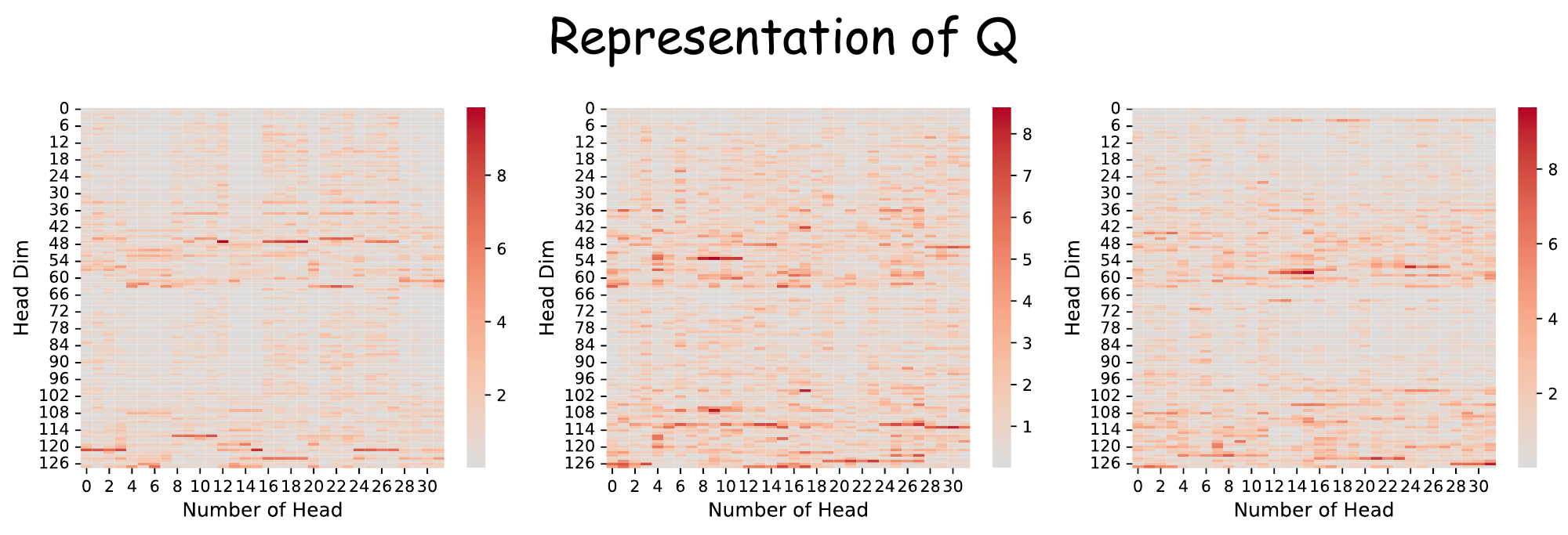}
    \caption{Visualization of the query representation magnitudes in Qwen3-8B on the RULER benchmark.}
    \label{fig_appendix:q}
\end{figure*}
\begin{figure*}
    \centering
    \includegraphics[width=0.8\linewidth]{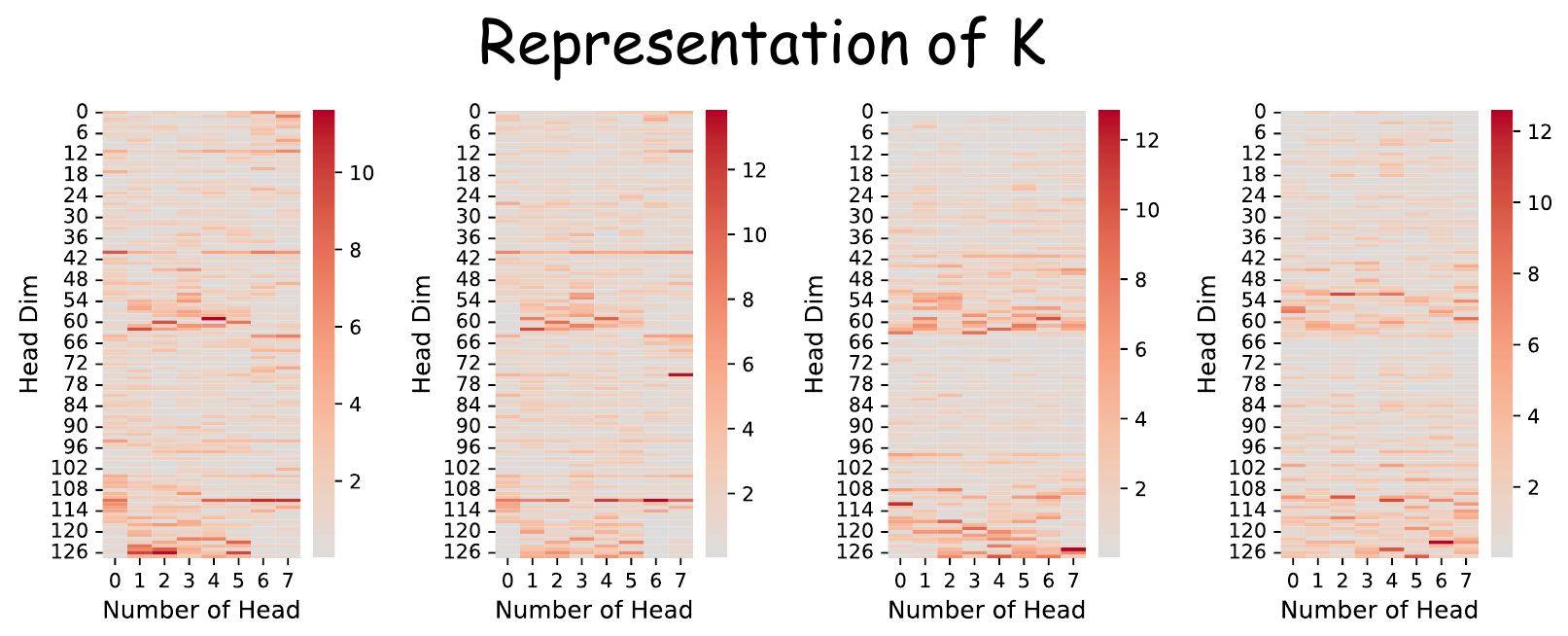}
    \caption{Visualization of the key representation magnitudes in Qwen3-8B on the RULER benchmark.}
    \label{fig_appendix:k}
\end{figure*}
\begin{figure*}
    \centering
    \includegraphics[width=0.8\linewidth]{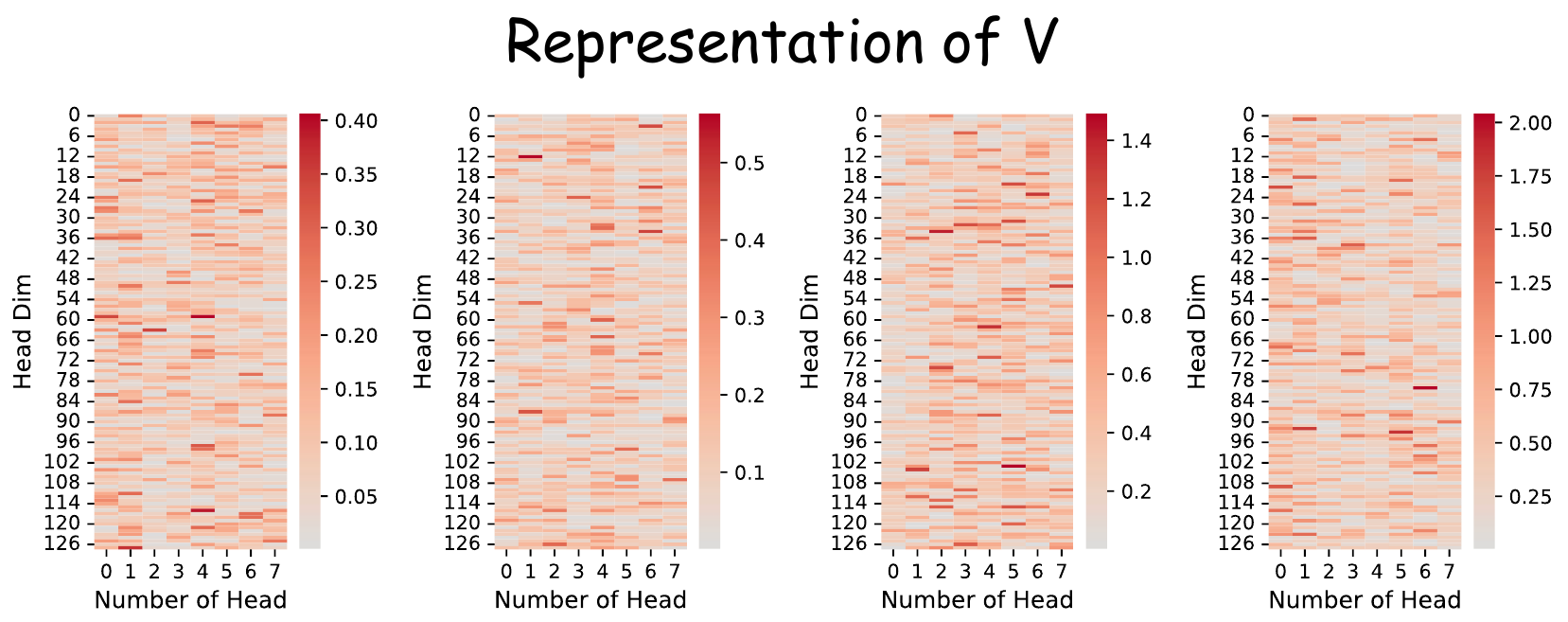}
    \caption{Visualization of the value representation magnitudes in Qwen3-8B on the RULER benchmark.}
    \label{fig_appendix:v}
\end{figure*}

\begin{figure*}
\centering
\begin{tcolorbox}[size=title,opacityfill=0.1]
\noindent
\faUser\ \textbf{Question}: Which of the following suggestions for the new year's initiatives is correct?
\\ \\
\faCogs\ \textbf{Qwen3-8B}:

<think>\textbackslash nOkay, let's tackle this question. The user is asking which of the given choices ……\textbackslash n\textbackslash n \textcolor[HTML]{2EA121}{First}, I need to recall…… \textcolor[HTML]{2EA121}{but} the option says looseness, which could be incorrect. Therefore, the correct answer should be D.\textbackslash n</think>\textbackslash n\textbackslash n\textcolor{red}{The correct answer is (D).}
\\ \\
\faCogs\ \textbf{Qwen3-8B w/o Non High-magnitude Activations}:

<think>\textbackslash nOkay, let's tackle this question. The user provided a text……\textbackslash n\textbackslash n\textcolor[HTML]{2EA121}{First}, I need to understand……\textbackslash n\textbackslash n\textcolor[HTML]{2EA121}{Alternatively}, the mention of coordination……\textcolor[HTML]{2EA121}{However}, the original text doesn't……So option D could be correct\textbackslash n</think>\textbackslash n\textbackslash n\textcolor{red}{The correct answer is (D)}.
\\

\faCogs\ \textbf{Qwen3-8B w/o High-magnitude Activations}:

\texttt{333333……3333333333333} 
\end{tcolorbox}
\caption{This is an example taken from the LongBench v2 dataset with the ID \texttt{66ec41d3821e116aacb1c874}, whose ground truth is \textbf{D}. The context is too long to be displayed in the image. "Qwen3-8B w/o High-magnitude Activations" refers to disrupting the top 30\% of high-magnitude activations, while "Qwen3-8B w/o Non High-magnitude Activations" refers to disrupting the smallest 30\% of activations. Numerous cases in LongBench v2 exhibit similar patterns.}
\label{case:massive_values}
\end{figure*}

\begin{table}[t]
\centering
\label{tab:massive_values_perturb}
\resizebox{1.0\linewidth}{!}{
    \begin{tabular}{l|c}
    \toprule
    \textbf{Setting} & \textbf{Correct / Total} \\
    \midrule
    \texttt{Qwen3-8B} & 173 / 503 \\
    \texttt{Qwen3-8B w/o Non High-magnitude Activations} & 108 / 503 \\
    \texttt{Qwen3-8B w/o High-magnitude Activations} & 0 / 503 \\
    \bottomrule
    \end{tabular}
}
\caption{Accuracy of Qwen3-8B on LongBench v2. We clamp selected activations to the global mean and report the number of correct predictions. }
\end{table}


\section{Discussion}
\label{app:discussion}

\paragraph{Sequence-Dimension Activation Analysis.}
We analyzed the activation magnitudes ($\ell_2$-norm) across the \emph{sequence} dimension of the Q and K representations. Our heatmap visualizations revealed no distinct or consistent high-magnitude patterns along the sequence dimension, in contrast to the clear patterns observed along the hidden-feature dimension (\cref{fig:1}). This finding is consistent with prior literature on activation-aware quantization~\citep{lin2024awq,liu2024kivi}, which consistently highlights massive activations within hidden dimensions rather than the sequence dimension. This further supports our design choice to focus on feature-level saliency rather than token-level selection.

\paragraph{Theoretical Justification.}
We provide a theoretical motivation for LongAct from two complementary perspectives.
From a \emph{representation} perspective, recent work by \citet{jin2025massive} demonstrates that high-magnitude activations in self-attention modules encode \emph{contextual knowledge}---information grounded in the current input context rather than parametric knowledge stored in weights. Long-context tasks fundamentally rely on understanding such contextual information. As shown in \cref{case:massive_values}, masking high-magnitude activations completely collapses retrieval (0/503 correct), whereas masking low-magnitude activations preserves substantial performance (108/503 correct). By targeting these high-magnitude channels, LongAct precisely focuses optimization on the context-processing circuitry.
From an \emph{optimization} perspective, unlike supervised fine-tuning which induces dense parameter updates, RL fine-tuning exhibits a fundamentally different dynamic. Recent studies~\citep{mukherjee2025reinforcement} reveal that effective RL updates are intrinsically sparse, modifying only a small functional subnetwork (5\%--30\% of parameters) to align model behavior. By freezing the vast majority of non-essential parameters, LongAct naturally adheres to this sparse update regime, helping the optimizer converge to the robust subnetwork required for long-context reasoning.
Together, these perspectives suggest that LongAct can be understood as selectively fine-tuning the activation subnetwork responsible for contextual understanding, while adhering to the sparse update regime essential for stable RL.

\paragraph{Comparison with MIGU.}
A related method, MIGU~\citep{du2024migu}, uses output magnitude (L1-norm) across different tasks to isolate task-specific parameters and mitigate catastrophic forgetting in multi-task settings. In contrast, LongAct targets a \emph{single-task} scenario where we observe intrinsic sparsity in activations (L2-norm) \emph{within} the task. We use this as a saliency guide to filter noisy gradient updates and enhance reasoning within the task (e.g., +7.63 on the Hard split of LongBench v2). Furthermore, LongAct reveals that these high-magnitude activations vary across different attention heads (\cref{fig:1}), providing a more fine-grained, head-aware update strategy.

\end{document}